\documentclass[10pt,twocolumn,letterpaper]{article}

\usepackage[pagenumbers]{cvpr} 

\usepackage{import}
\usepackage{bbm}
\usepackage{xcolor}
\usepackage{booktabs}
\usepackage{multirow}
\usepackage{pifont}
\usepackage[table,xcdraw]{xcolor}

\newcommand{\yus}{\scalebox{1.2}{\textcolor{green!60!black}{\ding{51}}}}
\newcommand{\nope}{\scalebox{1.2}{\textcolor{red}{\ding{55}}}}

\definecolor{cvprblue}{rgb}{0.21,0.49,0.74}
\usepackage[pagebackref,breaklinks,colorlinks,allcolors=cvprblue]{hyperref}

\def\paperID{The\_Latent\_Campus} 
\def\confName{MICCAI}
\def\confYear{2025}

\title{
Consistent View Alignment Improves Foundation Models \\ for 3D Medical Image Segmentation
}

\author{
Puru Vaish\textsuperscript{1}\thanks{Corresponding author} \quad
Felix Meister\textsuperscript{2} \quad
Tobias Heimann\textsuperscript{2} \quad
Christoph Brune \textsuperscript{1}\quad
Jelmer M. Wolterink \textsuperscript{1} \\
\textsuperscript{1} Department of Applied Mathematics, Technical Medical Centre, University of Twente \\
\textsuperscript{2} Digital Technology and Innovation, Siemens Healthineers, Erlangen, Germany \\
{\tt\small \{p.vaish, c.brune, j.m.wolterink\}@utwente.nl} \\
{\tt\small \{felix.meister, tobias.heimann\}@siemens-healthineers.com}
}

\begin{document}

\maketitle


\begin{abstract}
Many recent approaches in representation learning implicitly assume that uncorrelated views of a data point are sufficient to learn meaningful representations for various downstream tasks. In this work, we challenge this assumption and demonstrate that meaningful structure in the latent space does not emerge naturally. Instead, it must be explicitly induced. We propose a method that aligns representations from different views of the data to align complementary information without inducing false positives. 
Our experiments show that our proposed self-supervised learning method, \textit{Consistent View Alignment}, improves performance for downstream tasks, highlighting the critical role of structured view alignment in learning effective representations. Our method achieved first and second place in the MICCAI 2025 SSL3D challenge when using a Primus vision transformer and ResEnc convolutional neural network, respectively. The code and pretrained model weights are released at \href{https://github.com/Tenbatsu24/LatentCampus}{\texttt{github.com/Tenbatsu24/LatentCampus}}.

\end{abstract}    
\section{Introduction}
\label{sec:intro}

Learning effective and robust data representations is a central challenge in machine learning, with importance across supervised, semi-supervised, and self-supervised paradigms.
Contrastive learning frameworks have recently achieved remarkable success by training models to distinguish between similar (positive) and dissimilar (negative) sample pairs, capturing semantic patterns from both labelled and unlabelled data ~\cite{oord_representation_2018, chen_simple_2020, mathilde_caron_unsupervised_2020, caron_emerging_2021, chuang_robust_2022, maxime_oquab_dinov2_2023}.
In computer vision, natural language processing, and multimodal tasks, contrastive methods have been shown to produce rich, transferable embeddings ~\cite{chuang_debiased_2020, baevski_wav2vec_2020, robinson_contrastive_2021, venkataramanan_is_2024}.

\begin{figure}[!tbh]
  \centering
  \includegraphics[width=0.45\textwidth]{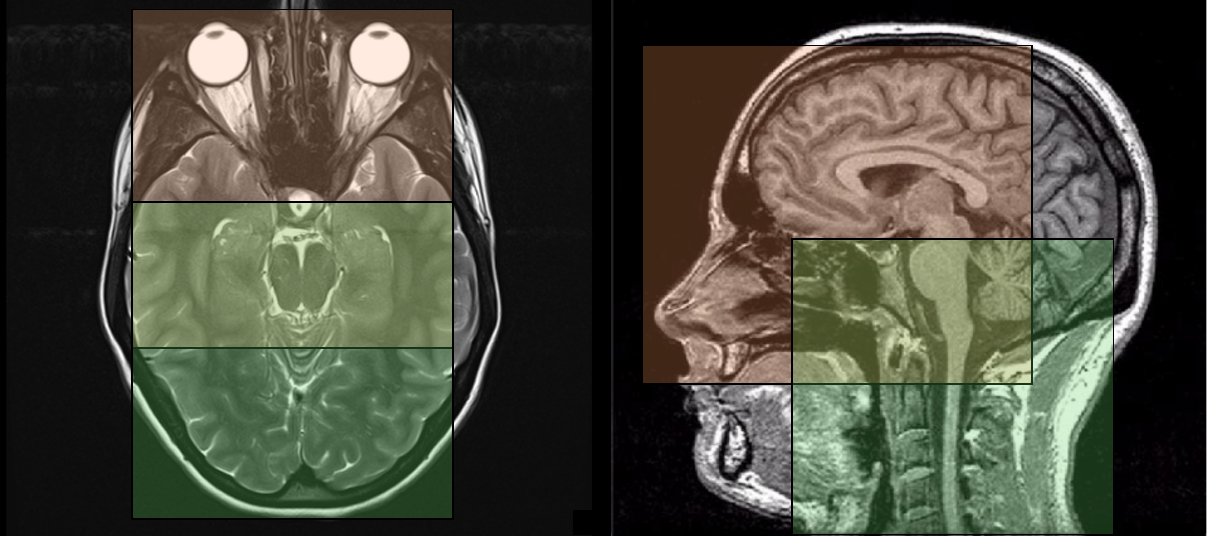}
   \caption{Two examples showcasing the issue of loosely correlated views. While two crops have an overlapping region between them, the two crops can be argued to represent completely different semantic information. However, self-supervised learning methods still treat them as positive pairs, forcing the model to align unrelated features, which can degrade representation quality for downstream tasks where ability to differentiate is important.}
   \label{fig:uncorrelated_views}
\end{figure}

However, contrastive learning relies critically on the assumption that positive pairs share meaningful, underlying information about the same instance~\cite{chuang_robust_2022}.
When this assumption is violated, such as through loosely correlated or mismatched views, models are forced to align unrelated regions, leading to spurious associations and ultimately degrading the quality of the learned representations~\cite{arjovsky_invariant_2019, jing_understanding_2022}.
While prior works have addressed related challenges via robust loss functions~\cite{ghosh_making_2015, wang_symmetric_2019}, view construction strategies~\cite{xiao_learning_2015, li_learning_2017}, and improved contrastive or mutual information estimators~\cite{ozair_wasserstein_2019}, these approaches primarily focus on pair selection or objective formulation.

Comparatively, little attention has been given to directly constraining where in the feature space alignment should occur.
In particular, current methods rarely enforce local, semantically consistent correspondences between views, leaving the latent space vulnerable to structural noise from false positives, as shown in Fig.~\ref{fig:uncorrelated_views}.
This motivates our central question: Can we explicitly regularise alignment in the feature space so that it occurs only between truly corresponding regions, thereby preserving meaningful structure and improving downstream generalisation?

To address this, we propose Consistent View Alignment (CVA), a novel regularisation technique that constructs and aligns views to mitigate the effects of false positives while enforcing local feature consistency. Our contributions are threefold:
\begin{itemize}
\item False positive mitigation via consistent view construction. We introduce a view alignment procedure that leverages explicit region correspondences to avoid uncorrelated or noisy positive pairs. By realigning feature targets at the patch or token level, CVA reduces the adverse impact of false positives that degrade representation quality in contrastive learning.
\item Structured representation learning for improved downstream tasks. We show that enforcing local feature consistency produces embeddings with stronger local consistency, leading to consistent improvements in downstream segmentation datasets across challenging benchmarks.
\item Feature space regularisation under varying spatial and semantic contexts. We regularise the latent space by enforcing consistency between matched regions across different spatial contexts, pixel/voxel details, and semantic content, operating at a fine-grained (patch/token) level rather than relying solely on global image or volume features.
\end{itemize}

By directly constraining where alignment occurs in the feature space, CVA addresses the core limitation of existing contrastive and reconstruction-based methods, preventing the collapse of representations due to false positives while preserving meaningful structure for robust downstream generalisation.





\section{Related Works}
\label{sec:related}

\subsection{Contrastive Learning in Natural Images}
Contrastive frameworks have set a strong foundation for representation learning by enforcing similarity between augmented views of the same instance while separating them from other instances. {SimCLR}~\cite{chen_simple_2020} remains a canonical starting point; its simplicity and effectiveness have inspired many later frameworks. However, its reliance on large batch sizes and strong positive-pair assumptions leaves it sensitive to false positives, a limitation that propagates to many descendants. {SwAV}~\cite{caron_emerging_2021} introduced \textit{loss symmetrisation} via bidirectional matching between view assignments, improving stability when fewer negatives are available. {SimSiam}~\cite{chen_exploring_2021} showed that collapse can be avoided without negative pairs, therefore being contrastive free, through a stop-gradient mechanism. However, it implicitly retains the same core dependence on clean, semantically aligned positive pairs. These methods set the stage for strong embeddings but offer little control over the \textit{structural regularity} of the learned feature space, which our work addresses directly.

\subsection{MAE and Hybrid Approaches}
{Masked Autoencoders (MAE)}~\cite{he_masked_2022} reconstruct heavily masked inputs from sparse visible patches, scaling exceptionally well without any explicit feature-space regularisation. Despite the absence of a contrastive signal, MAE produces highly transferable features across vision benchmarks. {Contrastive MAE (CMAE)}~\cite{huang_contrastive_2024} adds a contrastive component to MAE, further improving representation quality by encouraging view-invariant alignment in the latent space. However, CMAE still inherits the \textit{false positive sensitivity} of contrastive objectives and does not actively remove misaligned positives.

\subsection{3D Pretraining for Medical Images}
In 3D medical imaging, adapting pretraining to volumetric data presents unique challenges. {Volume Contrastive (VoCo)}~\cite{wu_voco_2024} extends contrastive learning to the 3D domain in a DINO-like fashion, using self-distillation with volume-specific augmentations. While effective for volumetric feature learning, its emphasis is on reducing \textit{false negatives} via more diverse positive sampling, leaving \textit{false positives} largely unaddressed. {Volume Fusion} approaches, such as {MIS-FM}~\cite{wang_mis-fm_2023}, fuse multiple volumes and frame the task as classifying which one differs from the rest. This implicitly becomes a boundary-finding problem rather than a representation learning task, limiting its ability to handle subtle semantic mismatches between volumes.

\subsection{Limitations and Our Focus}
From SimCLR’s foundational simplicity to MAE’s scalable masked reconstruction and VoCo’s volumetric self-distillation, these methods have made major strides in unsupervised and semi-supervised representation learning. Yet, across both 2D and 3D settings, one recurring gap remains: \textit{positive pairs are assumed to be perfectly correlated}. When this assumption fails, due to label noise, view mismatch, or inter-patient variability in medical scans, representations are forced to align unrelated data, degrading downstream performance. Our proposed \textit{Consistent View Alignment (CVA)} tackles this problem head-on by explicitly identifying and correcting false positives while regularising the local structure of the feature space, improving both robustness and generalisation.

\section{Methodology}\label{sec:method}

\begin{figure*}[tbh]
  \centering
  \def\svgwidth{\textwidth}
  \import{figures/schema}{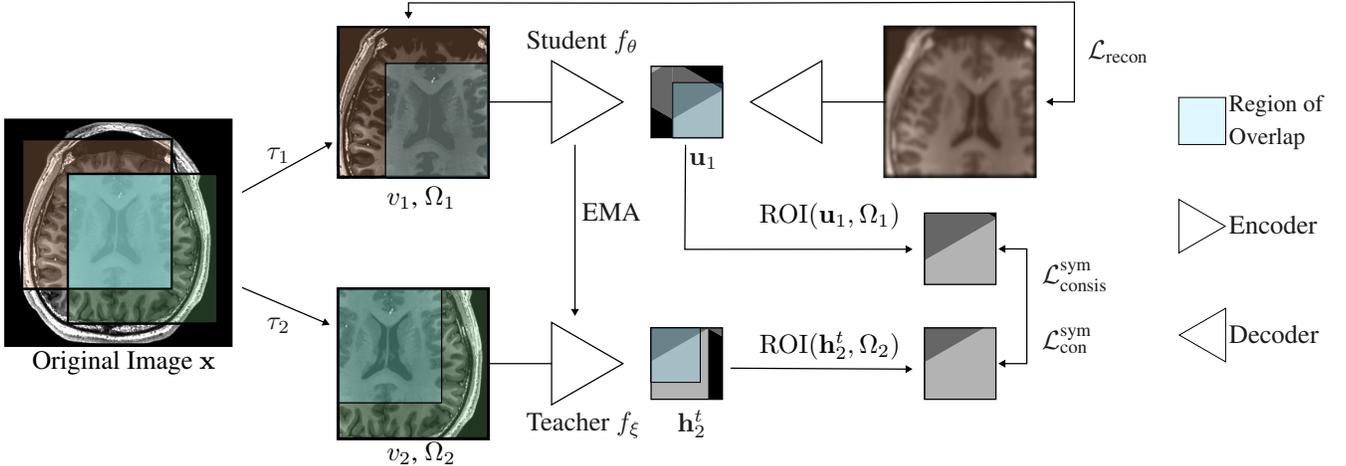}
   \caption{Overview of the proposed Consistent View Alignment (CVA) framework. Two random crops from the same image are produced such that they have an overlapping region (shaded blue) covering 40\% to 80\% of the image, ensuring a region of consistency between crops. They are then encoded by student and teacher networks (teacher updated via EMA). The reconstruction branch follows a masked autoencoding objective similar to standard reconstruction-based approaches, while the alignment branch uses bounding boxes of the overlapping regions to produce aligned feature maps via ROIAlign. A consistency loss enforces local feature agreement only on matched regions, mitigating SimCLR-like spurious alignments and regularising the latent space structure to yield consistent features across varying spatial contexts, pixel-level details, and semantic content.}
   \label{fig:method_in_images}
\end{figure*}

Our goal is to learn robust and transferable visual representations under scenarios where the standard positive-pair assumption in contrastive learning is violated due to uncorrelated views.
The Consistent View Alignment (CVA) framework that we propose is based on two individual parts, the first being view consistency and the second being feature alignment, both as shown in Fig.~\ref{fig:method_in_images}.

\subsection{Problem Formulation}
Let $\mathbf{x} \in \mathbb{R}^{C \times H \times W \times D}$ denote an input image or volume. We generate two random crops $v_1, v_2 = \mathcal{T}_1(\mathbf{x}), \mathcal{T}_2(\mathbf{x})$ via data augmentations $\mathcal{T}_1, \mathcal{T}_2$ drawn from a family $\mathcal{T}$. In conventional contrastive learning, $(v_1, v_2)$ are treated as a positive pair, assumed to represent the same underlying semantics. This assumption may not hold when crops cause a view mismatch or when the image contains heterogeneous content. Our approach mitigates this by sampling crops with a shared region, ensuring a region of consistency between crops.

\paragraph{Dual-Branch Architecture.}
We adopt a student-teacher architecture, where the \textbf{student encoder} $f_{\theta}$ is trained via gradient descent, and the \textbf{teacher encoder} $f_{\xi}$ is updated with an exponential moving average (EMA) of the student parameters:
\begin{equation}
\xi \leftarrow \tau \xi + (1 - \tau)\theta,
\end{equation}
where $\tau \in [0,1]$ controls the EMA decay rate. $\tau$ can be set to a non-zero value similar to BYOL~\cite{grill_bootstrap_2020}, DINO~\cite{caron_emerging_2021}, or the $\tau$ can be set to $0$, like in SimSiam~\cite{chen_exploring_2021}, where the teacher and student are always the same.

Given crops $v_1, v_2$, we obtain:
\begin{align}
\mathbf{z}_1^s &= f_{\theta}(v_1), \quad \mathbf{z}_2^s = f_{\theta}(v_2), \\
\mathbf{z}_1^t &= f_{\xi}(v_1), \quad \mathbf{z}_2^t = f_{\xi}(v_2),
\end{align}
where $\mathbf{z} \in \mathbb{R}^{\Tilde{C} \times \Tilde{H} \times \Tilde{W} \times \Tilde{D}}$ are spatial feature maps. 
Generally, $\Tilde{C} \gg C$ and the spatial dimensions are smaller.

\paragraph{Projector-Predictor Setup.} 
Following standard practice in contrastive learning~\cite{chen_simple_2020, caron_emerging_2021, huang_contrastive_2024}, the spatially pooled latent features $\mathbf{z} \in \mathbb{R}^{d}$ from the encoder are passed through a non-linear \textit{projector} $p_{\theta} : \mathbb{R}^{d} \rightarrow \mathbb{R}^{d_p}$, implemented as a two- or three-layer MLP:
\begin{equation}
\mathbf{h} = p_{\theta}(\mathbf{z}).
\end{equation}
For the student branch, the setup further appends a \textit{predictor} $q_{\theta} : \mathbb{R}^{d_p} \rightarrow \mathbb{R}^{d_p}$ that maps the projected features to the prediction space:
\begin{equation}
\mathbf{u} = q_{\theta}(\mathbf{h}),
\end{equation}
where $\mathbf{u}$ is compared to a target feature from another view, typically obtained from a (teacher) momentum encoder~\cite{grill_bootstrap_2020, huang_contrastive_2024} or a stop-gradient branch~\cite{chen_exploring_2021}. The intuition behind the predictor is to ``digest" information, while the projector attempts to disentangle information. This predictor-projector design has been shown to stabilise training and improve feature representation in self-distillation settings~\cite{chen_exploring_2021}.

\paragraph{Contrastive Loss with Symmetrisation.} 
The output of the projector-predictor can then be used in a contrastive learning objective similar to SimCLR~\cite{chen_simple_2020}, CMAE~\cite{huang_contrastive_2024}, with symmetrization like in swapped assignment from SwAV~\cite{caron_emerging_2021}. Let $\mathbf{u}_1$ and $\mathbf{u}_2$ denote the predictor outputs for two augmented views, and $\mathbf{h}^t_1$ and $\mathbf{h}^t_2$ the corresponding projected features from the target branch. The SimCLR loss can be rewritten as Normalized Temperature-scaled Cross Entropy (NT-Xent)~\cite{chen_intriguing_2021}, and in one direction it is:
\begin{equation}
\mathcal{L}_{\text{NT-Xent}}(\mathbf{u}_1, \mathbf{h}^t_2) = -\log \frac{\exp\left(\mathrm{sim}(\mathbf{u}_1, \mathbf{h}^t_2) / \tau\right)}{\sum\limits_{k=1}^{2N} \mathbbm{1}_{[k \neq i]} \exp\left(\mathrm{sim}(\mathbf{u}_1, \mathbf{h}^t_k) / \tau\right)},
\end{equation}
where $\mathrm{sim}(\cdot,\cdot)$ denotes cosine similarity, $\tau$ is a temperature parameter, and $N$ is the number of original samples in the batch (resulting in $2N$ augmented examples). The indicator $\mathbbm{1}_{[k \neq i]}$ masks out comparisons with the same view. This objective is \textit{symmetrised} by computing the loss in both directions and averaging:
\begin{equation}
\mathcal{L}_{\text{NT-Xent}}^{\text{sym}} = \frac{1}{2} \mathcal{L}_{\text{NT-Xent}}(\mathbf{u}_1, \mathbf{h}^t_2) + \frac{1}{2} \mathcal{L}_{\text{NT-Xent}}(\mathbf{u}_2, \mathbf{h}^t_1).
\end{equation}
In the SwAV formulation~\cite{mathilde_caron_unsupervised_2020}, the same principle applies in the assignment space: let $\mathbf{p}_1, \mathbf{p}_2$ be probability vectors over $K$ clusters obtained via Sinkhorn-Knopp normalisation, and $\mathbf{q}_1, \mathbf{q}_2$ the predicted assignments. The swapped prediction loss is:
\begin{equation}
\mathcal{L}_{\text{SwAV}} = -\frac{1}{2} \sum_{k=1}^K \mathbf{p}_{2,k} \log \mathbf{q}_{1,k} - \frac{1}{2} \sum_{k=1}^K \mathbf{p}_{1,k} \log \mathbf{q}_{2,k}.
\end{equation}
Symmetrisation has been shown to be effective in mitigating training instability and improving robustness to view and label noise~\cite{huang_contrastive_2024, yisen_wang_symmetric_2019, mathilde_caron_unsupervised_2020}.

\paragraph{Masked Autoencoding Reconstruction Loss.}
To encourage global semantic encoding, CMAE~\cite{he_masked_2022} introduces a masked reconstruction objective paired with a contrastive term, where the method randomly masks a high proportion $m$ of the volume/tokens \textbf{only} for the student and reconstructs the missing patches via a lightweight decoder $g_{\theta}$:
\begin{equation}
\mathcal{L}_{\text{recon}} = \frac{1}{|\mathcal{M}|} \sum_{i \in \mathcal{M}} l(g_{\theta}(\mathbf{z}_1^s)_i, v_{1,i}),
\end{equation}
where $\mathcal{M}$ indexes masked positions and $l$ is a specified distance function either $l_2$ or $l^\delta_{\text{huber}}$.

\begin{equation}
l_{\text{huber}}^{\delta}(y, \hat{y}) =
\begin{cases}
\frac{1}{2}(y - \hat{y})^2 & \text{if } |y - \hat{y}| \leq \delta \\
\delta \cdot \left( |y - \hat{y}| - \frac{1}{2}\delta \right) & \text{otherwise}
\end{cases}
\end{equation}

\begin{equation}
l_2(y, \hat{y}) = \frac{1}{n} \sum_{i=1}^{n} (y_i - \hat{y}_i)^2
\end{equation}

\subsection{Consistent View Alignment (CVA)}
The central novelty of CVA lies in an \textit{alignment step} that leverages spatial correspondences between two augmented views $v_1$ and $v_2$ to prevent false-positive alignments. Specifically, during the \textit{consistent view generation} step, we record the coordinates of the overlapping region between the two crops. This overlap is later used to spatially align feature maps before applying a consistency loss, ensuring that only semantically corresponding regions are compared.

\paragraph{Consistent View Generation.}
Given an original image $\mathbf{x} \in \mathbb{R}^{C \times H \times W \times D}$, we generate two crops $v_1, v_2 = \mathcal{T}_1(\mathbf{x}), \mathcal{T}_2(\mathbf{x})$ with a fixed output patch volume of $V_p$ using random spatial sampling with a constraint on overlap:
\begin{equation}
\gamma_{\text{min}} \leq \frac{\mathrm{Area}(\mathrm{Overlap}(v_1, v_2))}{V_p} \leq \gamma_{\text{max}}.
\end{equation}
Here $\gamma_{\text{min}} = 0.4$, $\gamma_{\text{max}} = 0.8$ and $V_p$ is set to volume of $160^3$ in our experiments. The resulting bounding boxes $\Omega_1$ and $\Omega_2$ representing the relative overlap location for each crop are stored for later use in feature alignment.

\paragraph{Feature Maps.}
For convolutional backbones, we follow VoCo~\cite{wu_voco_2024}. We extract feature maps from multiple stages, resize each to a fixed spatial resolution via bilinear interpolation, and concatenate them along the channel dimension:
\begin{align}
\mathbf{F} = \mathrm{Concat} \big[ &\mathrm{Resize}(\mathbf{F}_1), \dots, \mathrm{Resize}(\mathbf{F}_S) \big] \nonumber \\
&\in \mathbb{R}^{B \times C' \times H' \times W' \times D'}.
\end{align}
For transformer-based backbones we take the final encoder output before the classification head and perform the inverse of the tokenisation step to produce a spatial feature map similar to the convolution case.
When combined with a masked reconstruction objective, we follow Contrastive MAE~\cite{huang_contrastive_2024} and introduce a separate \textit{feature decoder} and \textit{pixel decoder} to predict the masked regions.

\paragraph{Feature Consistency.}
In the alignment step for reinforcing feature consistency, the feature maps are passed through the projector-predictor modules and then use RoiAlign ($\operatorname{ROI}$) using respective $\Omega$ to produce aligned feature maps of the same resolution:
\begin{equation}
\mathbf{h^\Omega} = \operatorname{ROI}(p_{\theta}(\mathbf{F}), \Omega), \quad
\mathbf{u^\Omega} = \operatorname{ROI}(q_{\theta}(\mathbf{h}), \Omega),
\end{equation}
where $p_{\theta}$ is a non-linear projector and $q_{\theta}$ a predictor network as introduced before.

Note that our feature maps are volumetric; therefore, we flatten the spatial dimensions after the alignment step. Now we can adopt the cosine regression loss from SimSiam~\cite{chen_exploring_2021}, which does not require negative examples:
\begin{equation}
\mathcal{L}_{\text{cos}}(\mathbf{u}^{\Omega_1}_1, \mathbf{h}^{{\Omega_2}, t}_2) = 2 - 2 \frac{\mathbf{u}^{\Omega_1}_1 \cdot \mathbf{h}^{{\Omega_2}, t}_2}{\|\mathbf{u}^{\Omega_1}_1\|_2 \; \|\mathbf{h}^{{\Omega_2}, t}_2\|_2}.
\end{equation}
We further adopt \textit{symmetrisation}~\cite{mathilde_caron_unsupervised_2020, huang_contrastive_2024} to compute the loss in both view directions:
\begin{equation}
\mathcal{L}^{\text{sym}}_{\text{cos}} = \frac{1}{2} \mathcal{L}_{\text{cos}}(\mathbf{u}^{\Omega_1}_1, \mathbf{h}^{{\Omega_2}, t}_2) + \frac{1}{2} \mathcal{L}_{\text{cos}}(\mathbf{u}^{\Omega_2}_2, \mathbf{h}^{{\Omega_1}, t}_1).
\end{equation}

\subsection{Overall Objective}
The overall training loss is a weighted combination of reconstruction, consistency, and optionally contrastive objectives:
\begin{equation}
\mathcal{L} = \lambda_{\text{recon}} \, \mathcal{L}_{\text{recon}} 
+ \lambda_{\text{consis}} \, \mathcal{L}^{\text{sym}}_{\text{consis}} 
+ \lambda_{\text{con}} \, \mathcal{L}^{\text{sym}}_{\text{con}},
\end{equation}
where $\lambda_{\text{recon}}$, $\lambda_{\text{consis}}$, and $\lambda_{\text{con}}$ are balancing hyperparameters. 

The reconstruction loss $\mathcal{L}_{\text{recon}}$ follows standard masked autoencoding loss, here it is fixed to the $l_{\text{huber}}$ loss. The consistency loss $\mathcal{L}^{\text{sym}}_{\text{consis}}$ enforces alignment between semantically corresponding regions, thereby reducing the impact of false positives and improving robustness. For this term, we can use either the symmetrised cosine regression loss ($\mathcal{L}^{\text{sym}}_{\text{cos}}$) or the symmetrised NT-Xent variant ($\mathcal{L}^{\text{sym}}_{\text{NT-Xent}}$), referred to as CVA and C-CVA respectively. Finally, the optional contrastive term $\mathcal{L}^{\text{sym}}_{\text{con}}$ further regularises the latent space to encourage transferable representations. 

Here, the superscript $\text{sym}$ indicates that we compute the losses in both view directions, following the symmetrisation scheme described in the previous section.



\section{Experiments and Results}
\label{sec:res}

\subsection{Experimental Setup}
\paragraph{Pre-training Dataset.}
We pretrain all models on the \textit{OpenMind} dataset~\cite{wald_openmind_2025}, the largest publicly available head \& neck MRI collection, comprising $114{,}570$ 3D volumes from $34{,}191$ patients. The dataset aggregates over $800$ smaller datasets released under CC0 or PDDL licences, harmonized in image format and metadata. It includes a wide spectrum of MRI sequences such as T1-weighted (T1w), T2-weighted (T2w), FLAIR, Mean Diffusion (MD), Fractional Anisotropy (FA), and several less common modalities.

\paragraph{Architectures.}
Note that our approach is, to some extent, agnostic to the kind of deep learning architecture used. We evaluate two representative architectures: the convolutional ResEnc-L~\cite{isensee_nnu-net_2024} and the transformer-based Primus-M~\cite{wald_primus_2025}. This allows us to test CVA across distinct inductive biases and computational trade-offs. Note Primus-M does not have a \texttt{cls} token and therefore the image latent vector is calculated by performing attention pooling~\cite{hassani_escaping_2021} instead of a simple average.

\paragraph{Augmentation Pipeline.}
Our augmentation strategy follows a spatial-then-color paradigm. First, spatial transformations (mirroring, scaling, rotation, simulated low resolution) are applied. Second, crop pairs with $40\%$-$80\%$ overlap are generated. Finally, an intensity transformation is applied from a set of noise injections (Rician or Gaussian), Gaussian blur ($\sigma\in\{0.5,1.0\}$), and photometric changes (brightness, contrast, gamma). While each crop can have a spatial transform independently, it complicates the alignment step, and hence this design choice was made. Validation uses the same consistent view generation but omits other transformations.

\paragraph{Training Specifications.}
We adopt a two-stage pretraining protocol. In \textit{stage one}, we train an MAE~\cite{he_masked_2022} baseline for $1000$ epochs: ResEnc-L ($\sim$5 days) and Primus-M ($\sim$14 days) on a single NVIDIA A40 GPU. These serve as initialization for all subsequent experiments, enabling shorter training cycles and reducing compute cost, particularly for transformer models. An MAE model was chosen as it appeared to be the pareto optimal model for segmentation and classification downstream tasks in 3D Brain Segmentation~\cite{wald_openmind_2025}. We fix the use of the reconstruction loss, $\mathcal{L}_{\text{rec}}$ to be the Huber reconstruction loss: $l^{\delta=1}_{\text{huber}}$.

In \textit{stage two} (\textit{post-pretraining}), we train all models with their respective objectives: ResEnc-L for $250$ epochs ($\sim$25 h, initial LR=$0.005$), and Primus-M for $50$ epochs ($\sim$48 h, initial LR=$0.0003$). We always use the final model for downstream fine-tuning tasks.

\paragraph{Downstream Fine-tuning.}
We evaluate on both segmentation and classification benchmarks:

\begin{itemize}
    \item \textbf{Segmentation:} 
    \begin{enumerate}
        \item \textbf{YBM} - Yale Brain Metastasis: \textit{enhancing tumor}, \textit{non-enhancing tumor}, and \textit{oedema}.
        \item \textbf{GLI} - BraTS Post-Glioblastoma Recurrence: \textit{enhancing tumor}, \textit{necrotic core}, \textit{oedema}, and \textit{non-enhancing tumor}.
        \item \textbf{ISL} - ISLES 2022 Ischemic Stroke Lesions: \textit{stroke lesion}.
        \item \textbf{MSD} - Medical Segmentation Decathlon Brain Tumor Segmentation: \textit{whole tumor}, \textit{tumor core}, and \textit{enhancing tumor}.
    \end{enumerate}
    For all segmentation datasets, we use a $50\%$/$50\%$ train-validation split, and report the mean of the average Dice Similarity Coefficient (DSC) and Normalized Surface Dice (NSD, 1\,mm threshold) computed per class.
    
    \item \textbf{Classification:} 
    \textbf{ABD II} - ABIDE-II (Autism Brain Imaging Data Exchange II), MPRAGE T1-weighted channel only: binary classification between \textit{Autism Spectrum Disorder (ASD)} and \textit{healthy control}. We report balanced accuracy, AUROC, and average precision, averaged over three-fold splits.
\end{itemize}
All fine-tuning runs last $150$ epochs with configurations detailed in our code repository.

\paragraph{Baselines and Ablations}
We compare our CVA method against: (1) MAE-trained baselines without post-pretraining, and (2) Contrastive MAE. Ablations investigate the effect of different consistency loss formulations and the addition of a contrastive term from Contrastive MAE in our framework.

\subsection{Results}

\begin{table*}[!tbh]
    \centering
    \caption{
Comparison of segmentation and classification performance across ResEnc-L and Primus-M tracks under different reconstruction, consistency, and contrastive configurations. 
The \textit{Recon.} column indicates the reconstruction strategy: \textbf{MAE} denotes Masked Autoencoder reconstruction, while \textbf{AE} corresponds to a standard Autoencoder without masking. 
The \textit{Consis.} column specifies the consistency regularization: a \nope~indicates no consistency loss; \textbf{CVA} refers to the objective using cosine similarity; \textbf{C-CVA} refers to the NT-Xent contrastive consistency objective.
The \textit{Cont.} column indicates whether a global contrastive term is used, in addition to reconstruction and consistency losses. 
Performance is reported for four segmentation datasets and one classification task. Lower ranks indicate better performance.
}
    \label{tab:results}
    \resizebox{\textwidth}{!}{
        \begin{tabular}{@{}clllllllllllllllll@{}}
\toprule
& \multicolumn{1}{c}{}                        & \multicolumn{1}{c}{}                         & \multicolumn{1}{c}{}                       & \multicolumn{1}{c}{}                          & \multicolumn{1}{c}{}                          & \multicolumn{1}{c}{}                          & \multicolumn{8}{c}{Segmentation}                                                                                                                                                                              & \multicolumn{3}{c}{Classification}                                                \\ \cmidrule(lr){8-15} \cmidrule(l){16-18} 
\multirow{2}{*}{Track}    & \multicolumn{1}{c}{\multirow{2}{*}{Recon.}} & \multicolumn{1}{c}{\multirow{2}{*}{Consis.}} & \multicolumn{1}{c}{\multirow{2}{*}{Cont.}} & \multicolumn{1}{c}{\multirow{2}{*}{Avg Rank}} & \multicolumn{1}{c}{\multirow{2}{*}{Seg Rank}} & \multicolumn{1}{c}{\multirow{2}{*}{Cls Rank}} & \multicolumn{2}{c}{ISL}                           & \multicolumn{2}{c}{YBM}                           & \multicolumn{2}{c}{GLI}                           & \multicolumn{2}{c}{MSD}                           & \multicolumn{3}{c}{ABD II}                                                        \\ \cmidrule(lr){8-9} \cmidrule(lr){10-11} \cmidrule(lr){12-13} \cmidrule(lr){14-15} \cmidrule(l){16-18} 
& \multicolumn{1}{c}{}                        & \multicolumn{1}{c}{}                         & \multicolumn{1}{c}{}                       & \multicolumn{1}{c}{}                          & \multicolumn{1}{c}{}                          & \multicolumn{1}{c}{}                          & \multicolumn{1}{c}{DSC} & \multicolumn{1}{c}{NSD} & \multicolumn{1}{c}{DSC} & \multicolumn{1}{c}{NSD} & \multicolumn{1}{c}{DSC} & \multicolumn{1}{c}{NSD} & \multicolumn{1}{c}{DSC} & \multicolumn{1}{c}{NSD} & \multicolumn{1}{c}{Bal Acc.} & \multicolumn{1}{c}{AUROC} & \multicolumn{1}{c}{AP} \\ \midrule
                           & AE                                           & \nope                                         & \nope                                       & 6.08                                           & 6.63                                           & 5.00                                           & \cellcolor[HTML]{FEDD81}77.34 & \cellcolor[HTML]{FEDB81}75.57 & \cellcolor[HTML]{C9DC81}60.92 & \cellcolor[HTML]{D3DF82}69.44 & \cellcolor[HTML]{FEE282}68.38 & \cellcolor[HTML]{FEE282}73.41 & \cellcolor[HTML]{9CCF7F}72.66 & \cellcolor[HTML]{E8E583}76.64 & \cellcolor[HTML]{F2E884}57.30 & \cellcolor[HTML]{F5E884}60.61 & \cellcolor[HTML]{CCDD82}60.03 \\
                           & MAE                                          & \nope                                         & \nope                                       & 5.22                                           & 5.00                                           & 5.67                                           & \cellcolor[HTML]{BAD881}78.87 & \cellcolor[HTML]{DDE283}76.66 & \cellcolor[HTML]{BAD881}61.21 & \cellcolor[HTML]{FEE683}68.68 & \cellcolor[HTML]{7EC67D}69.83 & \cellcolor[HTML]{7BC57D}75.02 & \cellcolor[HTML]{FEE883}72.22 & \cellcolor[HTML]{FEEA83}76.49 & \cellcolor[HTML]{F1E784}57.33 & \cellcolor[HTML]{FEE983}60.18 & \cellcolor[HTML]{FAEA84}58.89 \\ \cmidrule{2-18}
                           &                                           & CVA                                           & \nope                                       & 3.83                                           & 4.25                                           & 3.00                                           & \cellcolor[HTML]{FCEB84}77.98 & \cellcolor[HTML]{FFEB84}76.23 & \cellcolor[HTML]{94CD7E}62.10 & \cellcolor[HTML]{63BE7B}70.97 & \cellcolor[HTML]{E4E483}69.15 & \cellcolor[HTML]{D9E082}74.45 & \cellcolor[HTML]{6BC17C}72.84 & \cellcolor[HTML]{79C57D}77.15 & \cellcolor[HTML]{85C87D}60.14 & \cellcolor[HTML]{8BCA7E}63.69 & \cellcolor[HTML]{7FC67D}62.00 \\
                           & \multirow{-2}{*}{MAE}                                          & C-CVA                                         & \nope                                       & 5.14                                           & 4.38                                           & 6.67                                           & \cellcolor[HTML]{D0DE82}78.58 & \cellcolor[HTML]{D1DE82}76.81 & \cellcolor[HTML]{8ACA7E}62.27 & \cellcolor[HTML]{96CD7E}70.41 & \cellcolor[HTML]{ABD380}69.55 & \cellcolor[HTML]{B2D580}74.71 & \cellcolor[HTML]{AAD380}72.72 & \cellcolor[HTML]{B8D780}76.89 & \cellcolor[HTML]{FEE783}56.43 & \cellcolor[HTML]{FEE582}59.64 & \cellcolor[HTML]{FAEA84}59.06 \\ \cmidrule{2-18}
                           &                                           & \nope                                         & \yus                                        & 2.53                                           & 3.13                                           & \textbf{1.33}                                           & \cellcolor[HTML]{63BE7B}80.05 & \cellcolor[HTML]{66BF7C}78.18 & \cellcolor[HTML]{83C87D}62.31 & \cellcolor[HTML]{8BCA7E}70.37 & \cellcolor[HTML]{7FC67D}69.82 & \cellcolor[HTML]{8DCB7E}74.84 & \cellcolor[HTML]{75C37C}72.80 & \cellcolor[HTML]{DBE182}76.70 & \cellcolor[HTML]{63BE7B}61.09 & \cellcolor[HTML]{63BE7B}64.93 & \cellcolor[HTML]{63BE7B}62.67 \\
                           &                                           & CVA                                           & \yus                                        & \textbf{2.47}                                           & 2.88                                           & 1.67                                           & \cellcolor[HTML]{B3D580}78.97 & \cellcolor[HTML]{BBD881}77.09 & \cellcolor[HTML]{69C07C}62.35 & \cellcolor[HTML]{66BF7C}70.94 & \cellcolor[HTML]{81C77D}69.75 & \cellcolor[HTML]{97CD7E}74.85 & \cellcolor[HTML]{6CC17C}72.84 & \cellcolor[HTML]{99CE7F}77.02 & \cellcolor[HTML]{63BE7B}62.02 & \cellcolor[HTML]{72C37C}64.46 & \cellcolor[HTML]{65BF7C}62.62 \\
                           & \multirow{-3}{*}{MAE}                                          & C-CVA                                         & \yus                                        & 2.72                                           & \textbf{1.75}                                           & 4.67                                           & \cellcolor[HTML]{81C77D}79.65 & \cellcolor[HTML]{7BC57D}77.90 & \cellcolor[HTML]{7DC67D}62.43 & \cellcolor[HTML]{90CB7E}70.30 & \cellcolor[HTML]{71C27C}69.94 & \cellcolor[HTML]{6AC07C}75.18 & \cellcolor[HTML]{63BE7B}72.86 & \cellcolor[HTML]{63BE7B}77.24 & \cellcolor[HTML]{F7E984}57.17 & \cellcolor[HTML]{B6D680}62.48 & \cellcolor[HTML]{8ECB7E}61.60 \\ \cmidrule{2-18}
\multirow{-8}{*}{ResEnc-L} & \multicolumn{6}{c}{Range}                                                                                                                                                                                                                                                                     & 2.70                          & 2.61                          & 2.02                          & 2.28                          & 1.67                          & 1.83                          & 0.64                          & 0.74                          & 5.60                          & 5.28                          & 3.78                          \\ \midrule
                           & AE                                           & \nope                                         & \nope                                       & 5.03                                           & 6.38                                           & 2.33                                           & \cellcolor[HTML]{FEE582}76.05 & \cellcolor[HTML]{FED880}73.35 & \cellcolor[HTML]{DBE182}51.92 & \cellcolor[HTML]{FEE783}58.43 & \cellcolor[HTML]{FEDC81}63.35 & \cellcolor[HTML]{FEDB81}69.93 & \cellcolor[HTML]{F5E984}71.44 & \cellcolor[HTML]{96CD7E}75.90 & \cellcolor[HTML]{B9D780}56.09 & \cellcolor[HTML]{7EC67D}61.79 & \cellcolor[HTML]{79C57D}60.51 \\
                           & MAE                                          & \nope                                         & \nope                                       & 6.31                                           & 6.13                                           & 6.67                                           & \cellcolor[HTML]{ECE683}77.18 & \cellcolor[HTML]{CCDD82}74.98 & \cellcolor[HTML]{BCD881}52.70 & \cellcolor[HTML]{EAE583}59.01 & \cellcolor[HTML]{8CCA7E}65.82 & \cellcolor[HTML]{8CCA7E}72.58 & \cellcolor[HTML]{FEEA83}71.41 & \cellcolor[HTML]{FEE983}75.44 & \cellcolor[HTML]{E6E483}54.80 & \cellcolor[HTML]{FEEA83}58.75 & \cellcolor[HTML]{A7D27F}58.26 \\ \cmidrule{2-18}
                           &                                           & CVA                                           & \nope                                       & 4.64                                           & 4.63                                           & 4.67                                           & \cellcolor[HTML]{CADC81}77.18 & \cellcolor[HTML]{FEEB84}75.00 & \cellcolor[HTML]{ADD480}53.58 & \cellcolor[HTML]{C5DB81}59.95 & \cellcolor[HTML]{C5DB81}65.96 & \cellcolor[HTML]{C3DA81}72.73 & \cellcolor[HTML]{D2DE82}71.56 & \cellcolor[HTML]{F9EA84}75.54 & \cellcolor[HTML]{DBE182}55.83 & \cellcolor[HTML]{F0E784}59.17 & \cellcolor[HTML]{B0D580}58.38 \\
                           & \multirow{-2}{*}{MAE}                                          & C-CVA                                         & \nope                                       & 2.97                                           & 2.63                                           & 3.67                                           & \cellcolor[HTML]{EDE683}77.40 & \cellcolor[HTML]{DFE283}75.01 & \cellcolor[HTML]{B0D580}53.42 & \cellcolor[HTML]{DFE283}59.36 & \cellcolor[HTML]{63BE7B}67.21 & \cellcolor[HTML]{63BE7B}73.99 & \cellcolor[HTML]{71C27C}71.82 & \cellcolor[HTML]{63BE7B}76.14 & \cellcolor[HTML]{D4DF82}55.84 & \cellcolor[HTML]{ECE683}59.25 & \cellcolor[HTML]{9DCF7F}58.84 \\ \cmidrule{2-18}
                           &                                           & \nope                                         & \yus                                        & \textbf{2.50}                                           & 3.25                                           & \textbf{1.00}                                           & \cellcolor[HTML]{ECE683}77.18 & \cellcolor[HTML]{B5D680}75.36 & \cellcolor[HTML]{63BE7B}54.87 & \cellcolor[HTML]{63BE7B}61.74 & \cellcolor[HTML]{8DCA7E}65.82 & \cellcolor[HTML]{88C97E}72.65 & \cellcolor[HTML]{7FC67D}71.78 & \cellcolor[HTML]{A3D17F}75.85 & \cellcolor[HTML]{63BE7B}58.55 & \cellcolor[HTML]{63BE7B}62.42 & \cellcolor[HTML]{63BE7B}61.60 \\
                           &                                           & CVA                                           & \yus                                        & \textbf{2.49}                                           & \textbf{2.13}                                           & 3.21                                           & \cellcolor[HTML]{E9E583}77.33 & \cellcolor[HTML]{C3DA81}75.14 & \cellcolor[HTML]{67C07C}54.78 & \cellcolor[HTML]{6BC17C}61.60 & \cellcolor[HTML]{63BE7B}66.48 & \cellcolor[HTML]{63BE7B}73.27 & \cellcolor[HTML]{63BE7B}71.86 & \cellcolor[HTML]{63BE7B}76.10 & \cellcolor[HTML]{B8D780}56.13 & \cellcolor[HTML]{F2E884}59.10 & \cellcolor[HTML]{92CC7E}59.31 \\
                           & \multirow{-3}{*}{MAE}                                          & C-CVA                                         & \yus                                        & 4.03                                           & 2.88                                           & 6.33                                           & \cellcolor[HTML]{D9E082}78.07 & \cellcolor[HTML]{9BCF7F}75.77 & \cellcolor[HTML]{75C47D}54.44 & \cellcolor[HTML]{8CCA7E}60.92 & \cellcolor[HTML]{75C47D}66.20 & \cellcolor[HTML]{79C57D}72.91 & \cellcolor[HTML]{A9D27F}71.66 & \cellcolor[HTML]{DFE283}75.61 & \cellcolor[HTML]{CCDD82}55.55 & \cellcolor[HTML]{FCEB84}58.85 & \cellcolor[HTML]{B2D580}57.69 \\ \cmidrule{2-18}
\multirow{-8}{*}{Primus-M} & \multicolumn{6}{c}{Range}                                                                                                                                                                                                                                                                     & 2.01                          & 2.42                          & 2.95                          & 3.31                          & 3.86                          & 4.07                          & 0.45                          & 0.69                          & 3.75                          & 3.67                          & 3.90                          \\ \bottomrule
\end{tabular}

    }
\end{table*}

Table~\ref{tab:results} reports the performance of different reconstruction, consistency, and contrastive configurations across segmentation and classification tasks for both ResEnc-L and Primus-M tracks. We compare our proposed alignment-based consistency, CVA, which uses $\mathcal{L}_{\text{cos}}$ as the consistency loss, and C-CVA, which uses $\mathcal{L}_{\text{NT-Xent}}$, against MAE and Contrastive MAE baselines. Furthermore, we ablate on the combination and effect of the global contrastive signal in combination with our consistency loss. Lower ranks indicate stronger performance. To balance task importance, the overall score is defined as $2/3$ segmentation rank and $1/3$ classification rank. 

\paragraph{Segmentation.}
Alignment-based consistency (CVA/C-CVA) consistently improves segmentation performance across both tracks. In ResEnc-L, alignment variants achieve segmentation ranks of $4.25$/$4.38$, surpassing MAE ($5.00$). Furthermore, when combined with a contrastive loss, both models rank ($2.88$/$1.75$), outperforming Contrastive MAE ($3.13$). 
These gains are reflected in Dice, and NSD suggests that enforcing local feature consistency aids dense prediction. It is, however, interesting that both CVA and C-CVA benefit substantially from the existence of a contrastive term. 
The best performing segmentation ranking $1.75$, combines C-CVA with an additional contrastive term, therefore has two loss terms which are both contrastive in nature. This can likely explain better segmentation performance as the model is promoted to learn distinct fine-grained representations for smaller patches, which is absent in a non-contrastive consistency loss.
For Primus-M, alignment (CVA or C-CVA) alone yields the better segmentation rank ($4.63$/$2.63$), surpassing MAE ($6.13$) and C-CVA without a contrastive term actually outperforming Contrastive MAE ($3.25$) as well.
However, unlike ResEnc-L, the C-CVA alignment does not further improve when combined with the contrastive term. This indicates that the choice of alignment loss interacts with architecture, showcasing architecture-specific differences.  We get the best segmentation rank of $2.13$ when combining CVA with the contrastive term.

\paragraph{Classification.}
On ABIDE II, Contrastive MAE is the strongest performer in both tracks, achieving a rank of $1.33$ for ResEnc-L and $1.00$ for Primus-M outperforming alignment-based consistency. Although CVA and C-CVA remain competitive, their focus on local consistency appears to trade off against learning global, class-discriminative representations. While introducing the contrastive term generally boosts classification performance for the CVA consistency method, it cannot surpass the combination of MAE and contrastive loss alone. A combination of contrastive terms with a contrastive-based consistency (C-CVA) loss is always the worst performing classification model, even worse than a simple Auto Encoder trained model, highlighting that being dually contrastive conflicts with training a distinct single volume descriptor and learns poorly generalizing features. This highlights a key tension: while contrastive free alignment (CVA) benefits segmentation by refining spatial features, fully contrastive objectives capture higher-level semantic structure better suited for classification.

\paragraph{Overall Trends.}
The results illustrate a clear trade-off. Alignment-based consistency promotes robust segmentation by enforcing local spatial coherence, while contrastive objectives favour global discriminative features that generalize well to classification. Within this spectrum, ResEnc-L benefits most from CVA combined with a contrastive term, showing second-best segmentation improvements without catastrophic losses in classification. For Primus-M, we notice a tie (CVA with contrastive term ranking $2.49$ and Contrastive MAE ranking $2.50$) with no clear winner, as one model performs exceedingly well in segmentation and the other in classification. We also see that for both architectures, the best overall performance is achieved when consistent view alignment is combined with the contrastive term.

\subsection{Discussion}
In this paper, we have shown that alignment and consistency primarily benefit segmentation tasks by sharpening local representations, whereas contrastive pre-training provides more stable gains in classification through enhanced global separability. Consistent View Alignment with contrastive signal, in particular, emerged as the most reliable method across tasks, while Contrastive MAE was most stable for classification across folds. These findings highlight the trade-offs between pre-training strategies and suggest that no single approach dominates across all tasks. 

While the ranking analysis provides a useful overview, several factors complicate interpretation. First, ABIDE II classification is notoriously difficult: training is unstable, and many models diverge, making consistent improvement challenging. The stable convergence of Contrastive MAE likely contributed to its superior average rank, suggesting that its apparent dominance may partly reflect training robustness rather than strictly better representations. Our conclusions are also limited by the use of a single classification dataset, and future work should consider additional datasets and potentially revised training protocols to better evaluate pretrained representations for classification tasks.

Second, rankings can distort differences in both directions. On MSD, for instance, the absolute performance range between the best and worst models is less than $0.75\%$, which is practically negligible. Yet the ranking system amplifies such tiny gaps into substantial score differences. Conversely, when the absolute performance gap is large, a rank-based view may compress the apparent margin and downplay meaningful distinctions. There are at least two possible explanations for the small dynamic range (assuming no annotation errors): (1) the dataset is so challenging that pre-training offers little benefit, yielding uniformly similar scores, or (2) the small but consistent gaps reflect genuine improvements in representation learning precisely because the dataset is difficult. This underscores the importance of interpreting ranks with caution, especially in datasets with limited sensitivity to pre-training. In Appendix~\ref{sec:alt_rankings}, we present an alternative ranking scheme that accounts for small dynamic ranges.

Finally, our results reinforce the intuition that different objectives offer complementary benefits. Alignment improves segmentation by encouraging locally coherent features, whereas contrastive pre-training enhances classification by emphasizing global separability, often at the expense of fine detail. While being dually contrastive (C-CVA) may improve segmentation, it substantially degrades classification performance. 
To illustrate the flexibility of our framework, we further explore an alternative Gram based consistency variant in Appendix~\ref{sec:gram_consis}, 
demonstrating how second-order alignment objectives can be incorporated. 
Future work should explore hybrid objectives or task-adaptive weighting schemes that adjust the strength of consistency depending on the downstream application.

\section{Conclusion}\label{sec:conc}

Our study introduced consistent view alignment (CVA) as a lightweight extension to masked autoencoding, designed to improve medical image representation learning. Through experiments across four segmentation datasets and one challenging classification benchmark, we demonstrated that CVA and its NT-Xent variant (C-CVA) consistently strengthen segmentation performance by enforcing local spatial consistency. When combined with a global contrastive term, alignment-based methods achieve the strongest segmentation ranks, highlighting their ability to refine fine-grained features crucial for dense prediction tasks.

At the same time, our results reveal a fundamental trade-off: alignment excels in segmentation by promoting local coherence, whereas contrastive pretraining remains the most stable and effective choice for classification, where global discriminative features dominate. Being “doubly contrastive” (C-CVA with an added contrastive term) proved detrimental to classification, suggesting that overemphasis on contrastive objectives can harm global generalization. This tension underscores the need for carefully balancing local alignment and global contrastive signals.

Overall, CVA provides a robust alternative to contrastive pretraining in settings where spatial detail is paramount, while contrastive MAE remains preferable for classification-heavy tasks. These findings motivate future work on hybrid or adaptive objectives that can dynamically trade off between local alignment and global discrimination. Beyond performance gains, such methods could help bridge the gap between specialized medical models and more general-purpose foundation models, ensuring that representation learning can flexibly adapt to the demands of diverse downstream applications. Moreover, alignment-based objectives may also support interpretability by encouraging models to attend to localized structures, which is particularly valuable in clinical workflows where spatial reasoning underpins expert decision-making.

This conclusion is further supported by our performance in two MICCAI 2025 challenges: 
\textit{FOMO60k} and \textit{OpenMind SSL3D}. 
In the FOMO60k challenge, our method achieved the best segmentation performance on the validation stage of the open track (final results pending). 
In the OpenMind SSL3D challenge, our approach ranked first across both tracks on the validation stage, evaluated on three segmentation and one classification dataset. 
In the final leaderboard, based on four private segmentation datasets and three classification datasets, we obtained 
2\textsuperscript{nd} place in the ResEnc-L track and 1\textsuperscript{st} place in the Primus track. 
For both challenges, we submitted the model trained with CVA including the contrastive signal.

\paragraph{Limitations and Future Work.}
Despite pretraining on the large-scale OpenMind dataset, the diversity of modalities and pathologies outside head \& neck MRI remains underexplored, raising questions about transferability to other anatomies. Classification performance, particularly on ABIDE II, also remains unstable and sensitive to training dynamics, calling for more systematic investigation into optimization stability and robustness. Our exploration of hybrid objectives was limited to simple additive combinations; future work should investigate more principled balancing strategies, such as adaptive loss weighting or curriculum-based schedules. We also expect alignment to benefit more from heterogeneous data sources (e.g., across anatomies or natural domains), where avoiding view noise is particularly valuable. Finally, while our two-stage method reduced the burden of training large foundational models, exploring resource-efficient transformer training (e.g., Primus-M) could further expand accessibility in real-world pipelines. More broadly, advancing resource-efficient pretraining will be essential for enabling smaller institutions and healthcare providers to adopt such methods, ensuring that the benefits of representation learning can reach beyond research settings and into everyday clinical practice.

{
    \small
    \bibliographystyle{ieeenat_fullname}
    \bibliography{main}

\begin{thebibliography}{29}
\providecommand{\natexlab}[1]{#1}
\providecommand{\url}[1]{\texttt{#1}}
\expandafter\ifx\csname urlstyle\endcsname\relax
  \providecommand{\doi}[1]{doi: #1}\else
  \providecommand{\doi}{doi: \begingroup \urlstyle{rm}\Url}\fi

\bibitem[Arjovsky et~al.(2019)Arjovsky, Bottou, Gulrajani, and Lopez-Paz]{arjovsky_invariant_2019}
Martín Arjovsky, Léon Bottou, Ishaan Gulrajani, and David Lopez-Paz.
\newblock Invariant risk minimization.
\newblock \emph{Corr}, abs/1907.2893, 2019.
\newblock arXiv: 1907.02893.

\bibitem[Baevski et~al.(2020)Baevski, Zhou, Mohamed, and Auli]{baevski_wav2vec_2020}
Alexei Baevski, Yuhao Zhou, Abdelrahman Mohamed, and Michael Auli.
\newblock wav2vec 2.0: a framework for self-supervised learning of speech representations.
\newblock In \emph{Advances in {Neural} {Information} {Processing} {Systems} 33: {Annual} {Conference} on {Neural} {Information} {Processing} {Systems} 2020, {Neurips} 2020, {December} 6-12, 2020, {Virtual}}, 2020.

\bibitem[Caron et~al.(2021)Caron, Touvron, Misra, Jégou, Mairal, Bojanowski, and Joulin]{caron_emerging_2021}
Mathilde Caron, Hugo Touvron, Ishan Misra, Hervé Jégou, Julien Mairal, Piotr Bojanowski, and Armand Joulin.
\newblock Emerging properties in self-supervised vision transformers.
\newblock In \emph{2021 {IEEE}/{CVF} {International} {Conference} on {Computer} {Vision}, {ICCV} 2021, {Montreal}, {QC}, {Canada}, {October} 10-17, 2021}, pages 9630--9640. IEEE, 2021.

\bibitem[Chen et~al.(2020)Chen, Kornblith, Norouzi, and Hinton]{chen_simple_2020}
Ting Chen, Simon Kornblith, Mohammad Norouzi, and Geoffrey~E. Hinton.
\newblock A simple framework for contrastive learning of visual representations.
\newblock In \emph{Proceedings of the 37th {International} {Conference} on {Machine} {Learning}, {ICML} 2020, 13-18 {July} 2020, {Virtual} {Event}}, pages 1597--1607. PMLR, 2020.

\bibitem[Chen et~al.(2021)Chen, Luo, and Li]{chen_intriguing_2021}
Ting Chen, Calvin Luo, and Lala Li.
\newblock Intriguing properties of contrastive losses.
\newblock In \emph{Proceedings of the 35th {International} {Conference} on {Neural} {Information} {Processing} {Systems}}, pages 11834--11845, Red Hook, NY, USA, 2021. Curran Associates Inc.

\bibitem[Chen and He(2021)]{chen_exploring_2021}
Xinlei Chen and Kaiming He.
\newblock Exploring simple siamese representation learning.
\newblock In \emph{2021 {IEEE}/{CVF} {Conference} on {Computer} {Vision} and {Pattern} {Recognition} ({CVPR})}, pages 15745--15753, Nashville, TN, USA, 2021. IEEE.

\bibitem[Chuang et~al.(2020)Chuang, Robinson, Lin, Torralba, and Jegelka]{chuang_debiased_2020}
Ching-Yao Chuang, Joshua Robinson, Yen-Chen Lin, Antonio Torralba, and Stefanie Jegelka.
\newblock Debiased contrastive learning.
\newblock In \emph{Advances in {Neural} {Information} {Processing} {Systems} 33: {Annual} {Conference} on {Neural} {Information} {Processing} {Systems} 2020, {Neurips} 2020, {December} 6-12, 2020, {Virtual}}, 2020.

\bibitem[Chuang et~al.(2022)Chuang, Hjelm, Wang, Vineet, Joshi, Torralba, Jegelka, and Song]{chuang_robust_2022}
Ching-Yao Chuang, R~Devon Hjelm, Xin Wang, Vibhav Vineet, Neel Joshi, Antonio Torralba, Stefanie Jegelka, and Yale Song.
\newblock Robust contrastive learning against noisy views.
\newblock In \emph{2022 {IEEE}/{CVF} {Conference} on {Computer} {Vision} and {Pattern} {Recognition} ({CVPR})}, pages 16649--16660, 2022.
\newblock ISSN: 2575-7075.

\bibitem[Ghosh et~al.(2015)Ghosh, Manwani, and Sastry]{ghosh_making_2015}
Aritra Ghosh, Naresh Manwani, and P.S. Sastry.
\newblock Making risk minimization tolerant to label noise.
\newblock \emph{Neurocomput.}, 160\penalty0 (C):\penalty0 93--107, 2015.

\bibitem[Grill et~al.(2020)Grill, Strub, Altché, Tallec, Richemond, Buchatskaya, Doersch, Pires, Guo, Azar, Piot, Kavukcuoglu, Munos, and Valko]{grill_bootstrap_2020}
Jean-Bastien Grill, Florian Strub, Florent Altché, Corentin Tallec, Pierre~H. Richemond, Elena Buchatskaya, Carl Doersch, Bernardo~Ávila Pires, Zhaohan Guo, Mohammad~Gheshlaghi Azar, Bilal Piot, Koray Kavukcuoglu, Rémi Munos, and Michal Valko.
\newblock Bootstrap your own latent - a new approach to self-supervised learning.
\newblock In \emph{Advances in {Neural} {Information} {Processing} {Systems} 33: {Annual} {Conference} on {Neural} {Information} {Processing} {Systems} 2020, {Neurips} 2020, {December} 6-12, 2020, {Virtual}}, 2020.

\bibitem[Hassani et~al.(2021)Hassani, Walton, Shah, Abuduweili, Li, and Shi]{hassani_escaping_2021}
Ali Hassani, Steven Walton, Nikhil Shah, Abulikemu Abuduweili, Jiachen Li, and Humphrey Shi.
\newblock Escaping the big data paradigm with compact transformers.
\newblock \emph{Corr}, abs/2104.5704, 2021.
\newblock arXiv: 2104.05704.

\bibitem[He et~al.(2022)He, Chen, Xie, Li, Dollar, and Girshick]{he_masked_2022}
Kaiming He, Xinlei Chen, Saining Xie, Yanghao Li, Piotr Dollar, and Ross Girshick.
\newblock Masked autoencoders are scalable vision learners.
\newblock In \emph{2022 {IEEE}/{CVF} {Conference} on {Computer} {Vision} and {Pattern} {Recognition} ({CVPR})}, pages 15979--15988, New Orleans, LA, USA, 2022. IEEE.

\bibitem[Huang et~al.(2024)Huang, Jin, Lu, Hou, Cheng, Fu, Shen, and Feng]{huang_contrastive_2024}
Zhicheng Huang, Xiaojie Jin, Chengze Lu, Qibin Hou, Ming-Ming Cheng, Dongmei Fu, Xiaohui Shen, and Jiashi Feng.
\newblock Contrastive masked autoencoders are stronger vision learners.
\newblock \emph{IEEE Transactions on Pattern Analysis and Machine Intelligence}, 46\penalty0 (4):\penalty0 2506--2517, 2024.

\bibitem[Isensee et~al.(2024)Isensee, Wald, Ulrich, Baumgartner, Roy, Maier-Hein, and Jäger]{isensee_nnu-net_2024}
Fabian Isensee, Tassilo Wald, Constantin Ulrich, Michael Baumgartner, Saikat Roy, Klaus Maier-Hein, and Paul~F. Jäger.
\newblock {nnU}-net revisited: a call for rigorous validation in {3D} medical image segmentation.
\newblock In \emph{Medical {Image} {Computing} and {Computer} {Assisted} {Intervention} – {MICCAI} 2024: 27th {International} {Conference}, {Marrakesh}, {Morocco}, {October} 6–10, 2024, {Proceedings}, {Part} {IX}}, pages 488--498, Berlin, Heidelberg, 2024. Springer-Verlag.

\bibitem[Jing et~al.(2022)Jing, Vincent, LeCun, and Tian]{jing_understanding_2022}
Li Jing, Pascal Vincent, Yann LeCun, and Yuandong Tian.
\newblock Understanding dimensional collapse in contrastive self-supervised learning.
\newblock In \emph{The {Tenth} {International} {Conference} on {Learning} {Representations}, {ICLR} 2022, {Virtual} {Event}, {April} 25-29, 2022}. OpenReview.net, 2022.

\bibitem[Li et~al.(2017)Li, Yang, Song, Cao, Luo, and Li]{li_learning_2017}
Yuncheng Li, Jianchao Yang, Yale Song, Liangliang Cao, Jiebo Luo, and Li-Jia Li.
\newblock Learning from noisy labels with distillation.
\newblock In \emph{2017 {IEEE} {International} {Conference} on {Computer} {Vision} ({ICCV})}, pages 1928--1936, 2017.
\newblock ISSN: 2380-7504.

\bibitem[{Mathilde Caron} et~al.(2020){Mathilde Caron}, Caron, {Ishan Misra}, Misra, {Julien Mairal}, Mairal, {Priya Goyal}, Goyal, {Piotr Bojanowski}, Bojanowski, {Armand Joulin}, and Joulin]{mathilde_caron_unsupervised_2020}
{Mathilde Caron}, Mathilde Caron, {Ishan Misra}, Ishan Misra, {Julien Mairal}, Julien Mairal, {Priya Goyal}, Priya Goyal, {Piotr Bojanowski}, Piotr Bojanowski, {Armand Joulin}, and Armand Joulin.
\newblock Unsupervised learning of visual features by contrasting cluster assignments.
\newblock \emph{Neural Information Processing Systems}, 2020.

\bibitem[{Maxime Oquab} et~al.(2023){Maxime Oquab}, {Timothée Darcet}, {Théo Moutakanni}, {Huy T. Vo}, {Marc Szafraniec}, {Vasil Khalidov}, {Pierre Fernandez}, {Daniel Haziza}, {Francisco Massa}, {Alaaeldin El-Nouby}, {Mahmoud Assran}, {Nicolas Ballas}, {Wojciech Galuba}, {Russell Howes}, {Po-Yao Huang}, {Shang-Wen Li}, {Ishan Misra}, {Michael Rabbat}, {Vasu Sharma}, {Gabriel Synnaeve}, {Huibin Xu}, {Hervé Jeǵou}, {Julien Mairal}, {Patrick Labatut}, {Armand Joulin}, and {Piotr Bojanowski}]{maxime_oquab_dinov2_2023}
{Maxime Oquab}, {Timothée Darcet}, {Théo Moutakanni}, {Huy T. Vo}, {Marc Szafraniec}, {Vasil Khalidov}, {Pierre Fernandez}, {Daniel Haziza}, {Francisco Massa}, {Alaaeldin El-Nouby}, {Mahmoud Assran}, {Nicolas Ballas}, {Wojciech Galuba}, {Russell Howes}, {Po-Yao Huang}, {Shang-Wen Li}, {Ishan Misra}, {Michael Rabbat}, {Vasu Sharma}, {Gabriel Synnaeve}, {Huibin Xu}, {Hervé Jeǵou}, {Julien Mairal}, {Patrick Labatut}, {Armand Joulin}, and {Piotr Bojanowski}.
\newblock {DINOv2}: learning robust visual features without supervision.
\newblock \emph{Trans. Mach. Learn. Res.}, 2023.
\newblock arXiv: 2304.07193.

\bibitem[Oord et~al.(2018)Oord, Li, and Vinyals]{oord_representation_2018}
Aäron van~den Oord, Yazhe Li, and Oriol Vinyals.
\newblock Representation learning with contrastive predictive coding.
\newblock \emph{Corr}, abs/1807.3748, 2018.
\newblock arXiv: 1807.03748.

\bibitem[Ozair et~al.(2019)Ozair, Lynch, Bengio, van~den Oord, Levine, and Sermanet]{ozair_wasserstein_2019}
Sherjil Ozair, Corey Lynch, Yoshua Bengio, Aäron van~den Oord, Sergey Levine, and Pierre Sermanet.
\newblock Wasserstein dependency measure for representation learning.
\newblock In \emph{Proceedings of the 33rd {International} {Conference} on {Neural} {Information} {Processing} {Systems}}, number 1398, pages 15604--15614. Curran Associates Inc., Red Hook, NY, USA, 2019.

\bibitem[Robinson et~al.(2021)Robinson, Chuang, Sra, and Jegelka]{robinson_contrastive_2021}
Joshua~David Robinson, Ching-Yao Chuang, Suvrit Sra, and Stefanie Jegelka.
\newblock Contrastive learning with hard negative samples.
\newblock In \emph{9th {International} {Conference} on {Learning} {Representations}, {ICLR} 2021, {Virtual} {Event}, {Austria}, {May} 3-7, 2021}. OpenReview.net, 2021.

\bibitem[Venkataramanan et~al.(2024)Venkataramanan, Rizve, Carreira, Asano, and Avrithis]{venkataramanan_is_2024}
Shashanka Venkataramanan, Mamshad~Nayeem Rizve, João Carreira, Yuki~M. Asano, and Yannis Avrithis.
\newblock Is {ImageNet} worth 1 video? {Learning} strong image encoders from 1 long unlabelled video.
\newblock In \emph{The {Twelfth} {International} {Conference} on {Learning} {Representations}, {ICLR} 2024, {Vienna}, {Austria}, {May} 7-11, 2024}. OpenReview.net, 2024.

\bibitem[Wald et~al.(2025{\natexlab{a}})Wald, Roy, Isensee, Ulrich, Ziegler, Trofimova, Stock, Baumgartner, Köhler, and Maier-Hein]{wald_primus_2025}
Tassilo Wald, Saikat Roy, Fabian Isensee, Constantin Ulrich, Sebastian Ziegler, Dasha Trofimova, Raphael Stock, Michael Baumgartner, Gregor Köhler, and Klaus Maier-Hein.
\newblock Primus: enforcing attention usage for {3D} medical image segmentation, 2025{\natexlab{a}}.
\newblock arXiv:2503.01835 [cs].

\bibitem[Wald et~al.(2025{\natexlab{b}})Wald, Ulrich, Suprijadi, Ziegler, Nohel, Peretzke, Köhler, and Maier-Hein]{wald_openmind_2025}
Tassilo Wald, Constantin Ulrich, Jonathan Suprijadi, Sebastian Ziegler, Michal Nohel, Robin Peretzke, Gregor Köhler, and Klaus~H. Maier-Hein.
\newblock An {OpenMind} for {3D} medical vision self-supervised learning, 2025{\natexlab{b}}.
\newblock arXiv:2412.17041 [cs].

\bibitem[Wang et~al.(2023)Wang, Wu, Luo, Liu, Li, and Zhang]{wang_mis-fm_2023}
Guotai Wang, Jianghao Wu, Xiangde Luo, Xinglong Liu, Kang Li, and Shaoting Zhang.
\newblock {MIS}-{FM}: {3D} medical image segmentation using foundation models pretrained on a large-scale unannotated dataset.
\newblock \emph{Corr}, abs/2306.16925, 2023.
\newblock arXiv: 2306.16925.

\bibitem[Wang et~al.(2019)Wang, Ma, Chen, Luo, Yi, and Bailey]{wang_symmetric_2019}
Yisen Wang, Xingjun Ma, Zaiyi Chen, Yuan Luo, Jinfeng Yi, and James Bailey.
\newblock Symmetric cross entropy for robust learning with noisy labels.
\newblock In \emph{2019 {IEEE}/{CVF} {International} {Conference} on {Computer} {Vision} ({ICCV})}, pages 322--330, 2019.
\newblock ISSN: 2380-7504.

\bibitem[Wu et~al.(2024)Wu, Zhuang, and Chen]{wu_voco_2024}
Linshan Wu, Jiaxin Zhuang, and Hao Chen.
\newblock {VoCo}: a simple-yet-effective volume contrastive learning framework for {3D} medical image analysis.
\newblock In \emph{2024 {IEEE}/{CVF} {Conference} on {Computer} {Vision} and {Pattern} {Recognition} ({CVPR})}, pages 22873--22882, Seattle, WA, USA, 2024. IEEE.

\bibitem[Xiao et~al.(2015)Xiao, Xia, Yang, Huang, and Wang]{xiao_learning_2015}
Tong Xiao, Tian Xia, Yi Yang, Chang Huang, and Xiaogang Wang.
\newblock Learning from massive noisy labeled data for image classification.
\newblock In \emph{2015 {IEEE} {Conference} on {Computer} {Vision} and {Pattern} {Recognition} ({CVPR})}, pages 2691--2699, 2015.
\newblock ISSN: 1063-6919.

\bibitem[{Yisen Wang} et~al.(2019){Yisen Wang}, Wang, {Xingjun Ma}, Ma, {Zaiyi Chen}, {Zaiyi Chen}, Chen, {Yuan Luo}, {Yuan Luo}, {Yuan Luo}, Luo, {Jinfeng Yi}, Yi, {James E. Bailey}, and Bailey]{yisen_wang_symmetric_2019}
{Yisen Wang}, Yisen Wang, {Xingjun Ma}, Xingjun Ma, {Zaiyi Chen}, {Zaiyi Chen}, Zaiyi Chen, {Yuan Luo}, {Yuan Luo}, {Yuan Luo}, Yuan Luo, {Jinfeng Yi}, Jinfeng Yi, {James E. Bailey}, and James Bailey.
\newblock Symmetric cross entropy for robust learning with noisy labels.
\newblock \emph{IEEE International Conference on Computer Vision}, pages 322--330, 2019.
\newblock ARXIV\_ID: 1908.06112 MAG ID: 2981873476 S2ID: 3ba8d3060731d64cd46d27e933cbdfb8b7853f4b.

\end{thebibliography}
}

\clearpage
\setcounter{page}{1}
\maketitlesupplementary

\section{Alternate Rankings}\label{sec:alt_rankings}

While the main results section reported ranks based on raw performance ordering, such rankings may exaggerate differences on datasets where the absolute performance range is very narrow (e.g., MSD with less than $0.75\%$ difference across models). To investigate this, we re-compute rankings by weighting each metric according to its dynamic range. This approach down-weights datasets where all models perform nearly equally, thereby providing a more calibrated comparison of models across heterogeneous tasks.

\begin{table*}[tbh]
    \centering
    \caption{
    Range-weighted rankings of segmentation and classification performance across ResEnc-L and Primus-M tracks. 
    Each metric is first rescaled into a $[1,3]$ interval for interpretability using the transformation 
    $\text{score} = 3 - 2 \cdot \frac{\text{metric} - \min}{\max - \min}$, 
    where $\min$ and $\max$ denote the best and worst values observed across models for that metric. 
    This ensures that models performing at the range maximum receive a score of $1$, those at the minimum receive a $3$, and intermediate performances are interpolated linearly. 
    The final ranking aggregates these rescaled scores across datasets, thereby reducing the exaggeration of small performance gaps (e.g., in MSD) while still penalizing larger differences. Lower values indicate better performance. The metrics are repeated from above.
    }

    \label{tab:dyn_results}
    \resizebox{\textwidth}{!}{
        \begin{tabular}{@{}clllllllllllllllll@{}}
\toprule
& \multicolumn{1}{c}{}                        & \multicolumn{1}{c}{}                         & \multicolumn{1}{c}{}                       & \multicolumn{1}{c}{}                          & \multicolumn{1}{c}{}                          & \multicolumn{1}{c}{}                          & \multicolumn{8}{c}{Segmentation}                                                                                                                                                                              & \multicolumn{3}{c}{Classification}                                                \\ \cmidrule(lr){8-15} \cmidrule(l){16-18} 
\multirow{2}{*}{Track}    & \multicolumn{1}{c}{\multirow{2}{*}{Recon.}} & \multicolumn{1}{c}{\multirow{2}{*}{Consis.}} & \multicolumn{1}{c}{\multirow{2}{*}{Cont.}} & \multicolumn{1}{c}{\multirow{2}{*}{Avg Rank}} & \multicolumn{1}{c}{\multirow{2}{*}{Seg Rank}} & \multicolumn{1}{c}{\multirow{2}{*}{Cls Rank}} & \multicolumn{2}{c}{ISL}                           & \multicolumn{2}{c}{YBM}                           & \multicolumn{2}{c}{GLI}                           & \multicolumn{2}{c}{MSD}                           & \multicolumn{3}{c}{ABD II}                                                        \\ \cmidrule(lr){8-9} \cmidrule(lr){10-11} \cmidrule(lr){12-13} \cmidrule(lr){14-15} \cmidrule(l){16-18} 
& \multicolumn{1}{c}{}                        & \multicolumn{1}{c}{}                         & \multicolumn{1}{c}{}                       & \multicolumn{1}{c}{}                          & \multicolumn{1}{c}{}                          & \multicolumn{1}{c}{}                          & \multicolumn{1}{c}{DSC} & \multicolumn{1}{c}{NSD} & \multicolumn{1}{c}{DSC} & \multicolumn{1}{c}{NSD} & \multicolumn{1}{c}{DSC} & \multicolumn{1}{c}{NSD} & \multicolumn{1}{c}{DSC} & \multicolumn{1}{c}{NSD} & \multicolumn{1}{c}{Bal Acc.} & \multicolumn{1}{c}{AUROC} & \multicolumn{1}{c}{AP} \\ \midrule
                           & AE                                           & \nope                                         & \nope                                       & 2.65                                           & 2.69                                           & 2.57                                           & \cellcolor[HTML]{FEDD81}77.34 & \cellcolor[HTML]{FEDB81}75.57 & \cellcolor[HTML]{C9DC81}60.92 & \cellcolor[HTML]{D3DF82}69.44 & \cellcolor[HTML]{FEE282}68.38 & \cellcolor[HTML]{FEE282}73.41 & \cellcolor[HTML]{9CCF7F}72.66 & \cellcolor[HTML]{E8E583}76.64 & \cellcolor[HTML]{F2E884}57.30 & \cellcolor[HTML]{F5E884}60.61 & \cellcolor[HTML]{CCDD82}60.03 \\
                           & MAE                                          & \nope                                         & \nope                                       & 2.44                                           & 2.25                                           & 2.83                                           & \cellcolor[HTML]{BAD881}78.87 & \cellcolor[HTML]{DDE283}76.66 & \cellcolor[HTML]{BAD881}61.21 & \cellcolor[HTML]{FEE683}68.68 & \cellcolor[HTML]{7EC67D}69.83 & \cellcolor[HTML]{7BC57D}75.02 & \cellcolor[HTML]{FEE883}72.22 & \cellcolor[HTML]{FEEA83}76.49 & \cellcolor[HTML]{F1E784}57.33 & \cellcolor[HTML]{FEE983}60.18 & \cellcolor[HTML]{FAEA84}58.89 \\ \cmidrule{2-18}
                           &                                          & CVA                                           & \nope                                       & 1.63                                           & 1.70                                           & 1.50                                           & \cellcolor[HTML]{FCEB84}77.98 & \cellcolor[HTML]{FFEB84}76.23 & \cellcolor[HTML]{94CD7E}62.10 & \cellcolor[HTML]{63BE7B}70.97 & \cellcolor[HTML]{E4E483}69.15 & \cellcolor[HTML]{D9E082}74.45 & \cellcolor[HTML]{6BC17C}72.84 & \cellcolor[HTML]{79C57D}77.15 & \cellcolor[HTML]{85C87D}60.14 & \cellcolor[HTML]{8BCA7E}63.69 & \cellcolor[HTML]{7FC67D}62.00 \\
                           & \multirow{-2}{*}{MAE}                                          & C-CVA                                         & \nope                                       & 2.09                                           & 1.65                                           & 2.97                                           & \cellcolor[HTML]{D0DE82}78.58 & \cellcolor[HTML]{D1DE82}76.81 & \cellcolor[HTML]{8ACA7E}62.27 & \cellcolor[HTML]{96CD7E}70.41 & \cellcolor[HTML]{ABD380}69.55 & \cellcolor[HTML]{B2D580}74.71 & \cellcolor[HTML]{AAD380}72.72 & \cellcolor[HTML]{B8D780}76.89 & \cellcolor[HTML]{FEE783}56.43 & \cellcolor[HTML]{FEE582}59.64 & \cellcolor[HTML]{FAEA84}59.06 \\ \cmidrule{2-18}
                           &                                          & \nope                                         & \yus                                        & \textbf{1.28}                                           & 1.36                                           & 1.11                                           & \cellcolor[HTML]{63BE7B}80.05 & \cellcolor[HTML]{66BF7C}78.18 & \cellcolor[HTML]{83C87D}62.31 & \cellcolor[HTML]{8BCA7E}70.37 & \cellcolor[HTML]{7FC67D}69.82 & \cellcolor[HTML]{8DCB7E}74.84 & \cellcolor[HTML]{75C37C}72.80 & \cellcolor[HTML]{DBE182}76.70 & \cellcolor[HTML]{63BE7B}61.09 & \cellcolor[HTML]{63BE7B}64.93 & \cellcolor[HTML]{63BE7B}62.67 \\
                           &                                          & CVA                                           & \yus                                        & \textbf{1.27}                                           & 1.38                                           & \textbf{1.07}                                           & \cellcolor[HTML]{C7DB81}78.69 & \cellcolor[HTML]{D2DE82}76.80 & \cellcolor[HTML]{63BE7B}62.94 & \cellcolor[HTML]{6CC17C}70.88 & \cellcolor[HTML]{63BE7B}70.05 & \cellcolor[HTML]{63BE7B}75.24 & \cellcolor[HTML]{BAD881}72.69 & \cellcolor[HTML]{E5E483}76.71 & \cellcolor[HTML]{63BE7B}62.02 & \cellcolor[HTML]{72C37C}64.46 & \cellcolor[HTML]{65BF7C}62.62 \\
                           & \multirow{-3}{*}{MAE}                                          & C-CVA                                         & \yus                                        & 1.45                                           & \textbf{1.14}                                           & 2.08                                           & \cellcolor[HTML]{81C77D}79.65 & \cellcolor[HTML]{7BC57D}77.90 & \cellcolor[HTML]{7DC67D}62.43 & \cellcolor[HTML]{90CB7E}70.30 & \cellcolor[HTML]{71C27C}69.94 & \cellcolor[HTML]{6AC07C}75.18 & \cellcolor[HTML]{63BE7B}72.86 & \cellcolor[HTML]{63BE7B}77.24 & \cellcolor[HTML]{F7E984}57.17 & \cellcolor[HTML]{B6D680}62.48 & \cellcolor[HTML]{8ECB7E}61.60 \\ \cmidrule{2-18}
\multirow{-8}{*}{ResEnc-L} & \multicolumn{6}{c}{Range}                                                                                                                                                                                                                                                                     & 2.70                          & 2.61                          & 2.02                          & 2.28                          & 1.67                          & 1.83                          & 0.64                          & 0.74                          & 5.60                          & 5.28                          & 3.78                          \\ \midrule
                           & AE                                           & \nope                                         & \nope                                       & 2.45                                           & 2.81                                           & 1.74                                           & \cellcolor[HTML]{FEE582}76.05 & \cellcolor[HTML]{FED880}73.35 & \cellcolor[HTML]{DBE182}51.92 & \cellcolor[HTML]{FEE783}58.43 & \cellcolor[HTML]{FEDC81}63.35 & \cellcolor[HTML]{FEDB81}69.93 & \cellcolor[HTML]{F5E984}71.44 & \cellcolor[HTML]{96CD7E}75.90 & \cellcolor[HTML]{B9D780}56.09 & \cellcolor[HTML]{7EC67D}61.79 & \cellcolor[HTML]{79C57D}60.51 \\
                           & MAE                                          & \nope                                         & \nope                                       & 2.47                                           & 2.26                                           & 2.90                                           & \cellcolor[HTML]{ECE683}77.18 & \cellcolor[HTML]{CCDD82}74.98 & \cellcolor[HTML]{BCD881}52.70 & \cellcolor[HTML]{EAE583}59.01 & \cellcolor[HTML]{8CCA7E}65.82 & \cellcolor[HTML]{8CCA7E}72.58 & \cellcolor[HTML]{FEEA83}71.41 & \cellcolor[HTML]{FEE983}75.44 & \cellcolor[HTML]{E6E483}54.80 & \cellcolor[HTML]{FEEA83}58.75 & \cellcolor[HTML]{A7D27F}58.26 \\ \cmidrule{2-18}
                           &                                          & CVA                                           & \nope                                       & 2.19                                           & 1.97                                           & 2.62                                           & \cellcolor[HTML]{CADC81}77.18 & \cellcolor[HTML]{FEEB84}75.00 & \cellcolor[HTML]{ADD480}53.58 & \cellcolor[HTML]{C5DB81}59.95 & \cellcolor[HTML]{C5DB81}65.96 & \cellcolor[HTML]{C3DA81}72.73 & \cellcolor[HTML]{D2DE82}71.56 & \cellcolor[HTML]{F9EA84}75.54 & \cellcolor[HTML]{DBE182}55.83 & \cellcolor[HTML]{F0E784}59.17 & \cellcolor[HTML]{B0D580}58.38 \\
                           & \multirow{-2}{*}{MAE}                                          & C-CVA                                         & \nope                                       & 1.83                                           & 1.48                                           & 2.53                                           & \cellcolor[HTML]{EDE683}77.40 & \cellcolor[HTML]{DFE283}75.01 & \cellcolor[HTML]{B0D580}53.42 & \cellcolor[HTML]{DFE283}59.36 & \cellcolor[HTML]{63BE7B}67.21 & \cellcolor[HTML]{63BE7B}73.99 & \cellcolor[HTML]{71C27C}71.82 & \cellcolor[HTML]{63BE7B}76.14 & \cellcolor[HTML]{D4DF82}55.84 & \cellcolor[HTML]{ECE683}59.25 & \cellcolor[HTML]{9DCF7F}58.84 \\ \cmidrule{2-18}
                           &                                          & \nope                                         & \yus                                        & \textbf{1.31}                                           & 1.47                                           & \textbf{1.00}                                           & \cellcolor[HTML]{ECE683}77.18 & \cellcolor[HTML]{B5D680}75.36 & \cellcolor[HTML]{63BE7B}54.87 & \cellcolor[HTML]{63BE7B}61.74 & \cellcolor[HTML]{8DCA7E}65.82 & \cellcolor[HTML]{88C97E}72.65 & \cellcolor[HTML]{7FC67D}71.78 & \cellcolor[HTML]{A3D17F}75.85 & \cellcolor[HTML]{63BE7B}58.55 & \cellcolor[HTML]{63BE7B}62.42 & \cellcolor[HTML]{63BE7B}61.60 \\
                           &                                          & CVA                                           & \yus                                        & 1.66                                           & \textbf{1.28}                                           & 2.43                                           & \cellcolor[HTML]{E9E583}77.33 & \cellcolor[HTML]{C3DA81}75.14 & \cellcolor[HTML]{67C07C}54.78 & \cellcolor[HTML]{6BC17C}61.60 & \cellcolor[HTML]{63BE7B}66.48 & \cellcolor[HTML]{63BE7B}73.27 & \cellcolor[HTML]{63BE7B}71.86 & \cellcolor[HTML]{63BE7B}76.10 & \cellcolor[HTML]{B8D780}56.13 & \cellcolor[HTML]{F2E884}59.10 & \cellcolor[HTML]{92CC7E}59.31 \\
                           & \multirow{-3}{*}{MAE}                                          & C-CVA                                         & \yus                                        & 1.97                                           & 1.53                                           & 2.85                                           & \cellcolor[HTML]{D9E082}78.07 & \cellcolor[HTML]{9BCF7F}75.77 & \cellcolor[HTML]{75C47D}54.44 & \cellcolor[HTML]{8CCA7E}60.92 & \cellcolor[HTML]{75C47D}66.20 & \cellcolor[HTML]{79C57D}72.91 & \cellcolor[HTML]{A9D27F}71.66 & \cellcolor[HTML]{DFE283}75.61 & \cellcolor[HTML]{CCDD82}55.55 & \cellcolor[HTML]{FCEB84}58.85 & \cellcolor[HTML]{B2D580}57.69 \\ \cmidrule{2-18}
\multirow{-8}{*}{Primus-M} & \multicolumn{6}{c}{Range}                                                                                                                                                                                                                                                                     & 2.01                          & 2.42                          & 2.95                          & 3.31                          & 3.86                          & 4.07                          & 0.45                          & 0.69                          & 3.75                          & 3.67                          & 3.90                          \\ \bottomrule
\end{tabular}

    }
\end{table*}

In the new ranking method shown in Table~\ref{tab:dyn_results} we see that in the average rankings Contrastive MAE is much closer to the CVA with the same contrastive term for ResEnc-L and for Primus-M it actually surpasses CVA with the contrastive term by large margin. The relative rankings do not change for segmentation tasks. This new ranking method therefore, highlights the importance of the choice of ranking and to critically consider the existence of small dynamic range of metrics on a give task.

\section{Gram Consistency Variant}\label{sec:gram_consis}

\begin{table}[bth]
    \centering
    \caption{Performance of the Gram consistency variant across segmentation and classification benchmarks. 
    Values are reported as percentages. MAE was used as the reconstruction loss, and no contrastive signal was added.}
    \label{tab:gram_results}
    \resizebox{0.5\textwidth}{!}{
    \begin{tabular}{@{}cllllllllllll@{}}
    \toprule
    \multirow{2}{*}{Track} & \multicolumn{2}{c}{ISL} & \multicolumn{2}{c}{YBM} & \multicolumn{2}{c}{GLI} & \multicolumn{2}{c}{MSD} & \multicolumn{3}{c}{ABIDE II} \\ 
    \cmidrule(lr){2-3} \cmidrule(lr){4-5} \cmidrule(lr){6-7} \cmidrule(lr){8-9} \cmidrule(l){10-12}
    & DSC & NSD & DSC & NSD & DSC & NSD & DSC & NSD & Bal. Acc. & AUROC & AP \\ 
    \midrule
    ResEnc-L & 79.30 & 77.57 & 62.78 & 71.12 & 69.84 & 74.84 & 72.96 & 77.04 & 56.95 & 59.74 & 58.76 \\
    Primus-M & 77.80 & 75.58 & 53.93 & 60.99 & 66.14 & 72.97 & 71.38 & 75.59 & 54.86 & 58.21 & 57.20 \\
    \bottomrule
    \end{tabular}
}
\end{table}

To illustrate the flexibility of our framework, we experimented with an alternative consistency objective based on Gram matrix anchoring. 
Here, local feature correlations are matched across views by minimizing the Frobenius norm between their Gram matrices.

Suppose an image is divided into $p$ patches/voxels, and the network produces $d$-dimensional features. 
Let $z_{1}^{s} \in \mathbb{R}^{p \times d}$ (student) and 
$z_{2}^{t} \in \mathbb{R}^{p \times d}$ (teacher) denote the $\ell_2$-normalized patch features for two crops of the same volume. 
These features are taken directly after our alignment step (we omit $\Omega$ in superscript for clarity) and without a projector-predictor module. 
The Gram anchoring loss is defined as
\begin{equation}
\mathcal{L}_{\text{Gram}} = 
\big\| z_{1}^{s} {z_{1}^{s}}^\top - z_{2}^{t} {z_{2}^{t}}^\top \big\|_F^2,
\end{equation}
and can be symmetrized as
\begin{equation}
\mathcal{L}_{\text{Gram}}^{\text{sym}} = 
\frac{1}{2} \mathcal{L}_{\text{Gram}}(z_{1}^{s}, z_{2}^{t}) +
\frac{1}{2} \mathcal{L}_{\text{Gram}}(z_{2}^{s}, z_{1}^{t}).
\end{equation}

Compared to cosine regression, which operates in \emph{first-order space} by aligning features directly, 
Gram anchoring acts in \emph{second-order space} by aligning the pairwise similarity structure between features. 
First-order alignment enforces that corresponding feature vectors point in similar directions, whereas second-order alignment allows features to move freely as long as their relative correlations are preserved.

Table~\ref{tab:gram_results} summarizes the results. All models were trained with MAE reconstruction and no contrastive signal.

Overall, Gram consistency appears beneficial for ResEnc-L, with consistent improvements across segmentation datasets. 
For Primus-M, however, the gains are inconsistent, suggesting that transformer-based models may benefit less from this 
form of second-order alignment in the medical domain. Notably, classification performance on ABIDE II decreases substantially in both 
architectures, indicating a trade-off that requires further investigation. In future work, we plan to ablate 
whether combining Gram consistency with a contrastive signal can recover classification performance, similar to 
our findings with cosine regression and NT-Xent.

\end{document}